\documentclass[10pt,twocolumn,letterpaper]{article}

\usepackage{cvpr}              % To produce the CAMERA-READY version
\usepackage{graphicx}
\usepackage{amsmath}
\usepackage{amssymb}
\usepackage{booktabs}
\usepackage[ruled,vlined]{algorithm2e}
\usepackage{algpseudocode}
\usepackage{stfloats} %custom

\usepackage[pagebackref,breaklinks,colorlinks]{hyperref}

\usepackage[capitalize]{cleveref}
\crefname{section}{Sec.}{Secs.}
\Crefname{section}{Section}{Sections}
\Crefname{table}{Table}{Tables}
\crefname{table}{Tab.}{Tabs.}

\usepackage{enumerate}% http://ctan.org/pkg/enumerate
\RequirePackage{color}

\def\cvprPaperID{10} % *** Enter the CVPR Paper ID here
\def\confName{CVPR}
\def\confYear{2022}

\begin{document}

%%%%%%%%% TITLE - PLEASE UPDATE
\title{Entropy-based Stability-Plasticity for Lifelong Learning}

% \author{%
% Vladimir Araujo\textsuperscript{1,2}\thanks{Equal contribution.}, Julio Hurtado\textsuperscript{2,3}\footnotemark[1], Alvaro Soto\textsuperscript{2}, Marie-Francine Moens\textsuperscript{1}
% \\
% \normalsize \small\affaddr{\textsuperscript{1}KU Leuven}, \small\affaddr{\textsuperscript{2}Pontificia Universidad Católica de Chile},
% \small\affaddr{\textsuperscript{3}University of Pisa}
% \\
% \small \email{vgaraujo@uc.cl, jahurtado@uc.cl, asoto@ing.puc.cl, sien.moens@kuleuven.be}
% }
\author{Vladimir Araujo\textsuperscript{1,2}\thanks{Equal contribution.}, Julio Hurtado\textsuperscript{2,3}\footnotemark[1], Alvaro Soto\textsuperscript{2}, Marie-Francine Moens\textsuperscript{1}\\
\textsuperscript{1}KU Leuven, \textsuperscript{2}Pontificia Universidad Católica de Chile, \textsuperscript{3}University of Pisa \\
{\tt\small vgaraujo@uc.cl, jahurtado@uc.cl, asoto@ing.puc.cl, sien.moens@kuleuven.be}
% For a paper whose authors are all at the same institution,
% omit the following lines up until the closing ``}''.
% Additional authors and addresses can be added with ``\and'',
% just like the second author.
% To save space, use either the email address or home page, not both
}

\maketitle

%%%%%%%%% ABSTRACT
\begin{abstract}
   The ability to continuously learn remains elusive for deep learning models. Unlike humans, models cannot accumulate knowledge in their weights when learning new tasks, mainly due to an excess of plasticity and the low incentive to reuse weights when training a new task. To address the stability-plasticity dilemma in neural networks, we propose a novel method called Entropy-based Stability-Plasticity (ESP). Our approach can decide dynamically how much each model layer should be modified via a plasticity factor. We incorporate branch layers and an entropy-based criterion into the model to find such factor. Our experiments in the domains of natural language and vision show the effectiveness of our approach in leveraging prior knowledge by reducing interference. Also, in some cases, it is possible to freeze layers during training leading to speed up in training.
\end{abstract}

%%%%%%%%% BODY TEXT
\section{Introduction}
\label{sec:intro}

Humans learn continuously throughout their lives, integrating new information to their knowledge to face new and changing environments. 
By contrast, artificial neural networks learn in a bounded environment, where the input distribution is assumed fixed. When the input distribution changes, the model must adapt its weights to perform correctly on the new task. Due to those modifications, the model overwrites previously learned patterns, creating interference between old and new tasks, causing a problem known as catastrophic forgetting \cite{mccloskey1989catastrophic, ratcliff1990connectionist}. This excessive plasticity in the model is part of the stability-plasticity dilemma \cite{10.3389/fpsyg.2013.00504, 10.1371/journal.pcbi.1006604}, which addresses the trade-off between modifying the parameters to learn a new task (plasticity) or keeping the parameters constant (stability) to avoid interference between tasks.

Several methods have been proposed to mitigate the stability-plasticity dilemma, focusing mainly on avoiding the catastrophic forgetting problem. Using different techniques to mitigate interference, these methods can be divided into two groups. The first group aims to restrict weight modifications by using regularization functions \cite{Kirkpatrick2017, 10.5555/3305890.3306093, aljundi2018memory} that minimize the modifications of key weight values. The second group uses gating functions \cite{serra2018overcoming, mallya2018piggyback, hurtado2021overcoming} to adaptively activate each weight depending on the context provided by the current task or input instance.

In this work, we follow the first group by proposing a model that aims to restrict weight modifications.
% Following the line of work that attempts to restrict weight modifications, 
We rely on evidence showing that the lower layers of a deep learning model capture general knowledge while the upper layers capture task-specific knowledge \cite{Zeiler2014, NIPS2014_375c7134, Kovaleva2019, lee2019elsa}. Under this premise, in the case of a lifelong learning scenario, a model should update its layer weights based on how general or specific these layers should be.
We propose the Entropy-based Stability-Plasticity (ESP) method, which relies on an entropy-based criterion to decide how much a model has to modify the weights in each of its layers.
Specifically, ESP augments each layer of an encoder with a branch layer that computes an entropy-based plasticity factor during the forward pass, and dynamically updates the layer weights based on these plasticity factors during the backward pass.
This way, when a new training example arrives, the model calculates how much we can update the weights of the model via a plasticity factor. We found in our experiments that in some cases, our method forces the model to freeze some layers, setting their gradients to zero, which encourages reusing past knowledge and reduces training time.

We demonstrate the effectiveness of our method experimentally by running a diverse set of experiments and comparing our results against well-known baselines. Unlike previous work in the field, we evaluated ESP on both, vision and natural language domains. 
The code is publicly available for further replicability and future research\footnote{\href{https://github.com/vgaraujov/ESP-CL}{https://github.com/vgaraujov/ESP-CL}}. 

\section{Related Work}

Previous methods have tackle the problem of Continual Learning (CL) using three main strategies. The first group of methods focus on limiting the plasticity of learning new tasks. The typical approach penalizes weight modifications or freezes a subset of the model. This can be achieved by adding weight regularizations \cite{Kirkpatrick2017,10.5555/3305890.3306093}, using masks to freeze parts of the model \cite{mallya2018piggyback, mallya2018packnet, hurtado2021overcoming}, or based those regularization on gradients behavior \cite{chaudhry2018efficient, saha2021gradient}. 

The second strategy is to use dynamic architectures by increase network capacity and adding extra parameters \cite{rusu2016progressive, ebrahimi2020adversarial}, or by finding new paths of relevant weights to solve each task \cite{fernando2017pathnet}, freezing used weights, and limiting learning of new tasks. Other approaches use different functions as components in the network, either Hypernetworks \cite{Oswald2020Continual}, Deep Artificial Neurons (DANs) \cite{camp2020continual}, Compositional Structures \cite{mendez2021lifelong, ostapenko2021continual}, or novel learning strategies \cite{hurtado2021optimizing}, so that network components can be more flexible when learning new tasks.

The third strategy is based on memory-based methods. This strategy mitigate catastrophic forgetting by inserting data from past tasks into the training process of new tasks, continuously re-training previous tasks \cite{10.1162/neco_a_01433}, either with raw samples \cite{rebuffi2017icarl, chaudhry2019continual,pmlr-v119-chrysakis20a}, or minimizing gradient interference \cite{lopez2017gradient, chaudhry2018efficient}. Later works such as \cite{lesort2019generative, shin2017continual} train generator functions (GANs) or autoencoders \cite{kemker2018fearnet} to generate elements from past distributions. They seek memory-efficiency by generating examples instead of saving real data. Similarly, \cite{iscen2020memory, caccia2020aqm} seek to be memory-efficient by saving feature vectors of instances from previous tasks, while learning a transformation from the feature space of past tasks to current ones. Other works use memory to create prototypes that can represent classes \cite{rebuffi2017icarl, de2021continual}, either for use as distillation or classification vectors.

\section{Method}\label{method}

\begin{figure}[t]
  \centering
%   \fbox{\rule{0pt}{2in} \rule{0.9\linewidth}{0pt}}
  \includegraphics[width=0.95\linewidth]{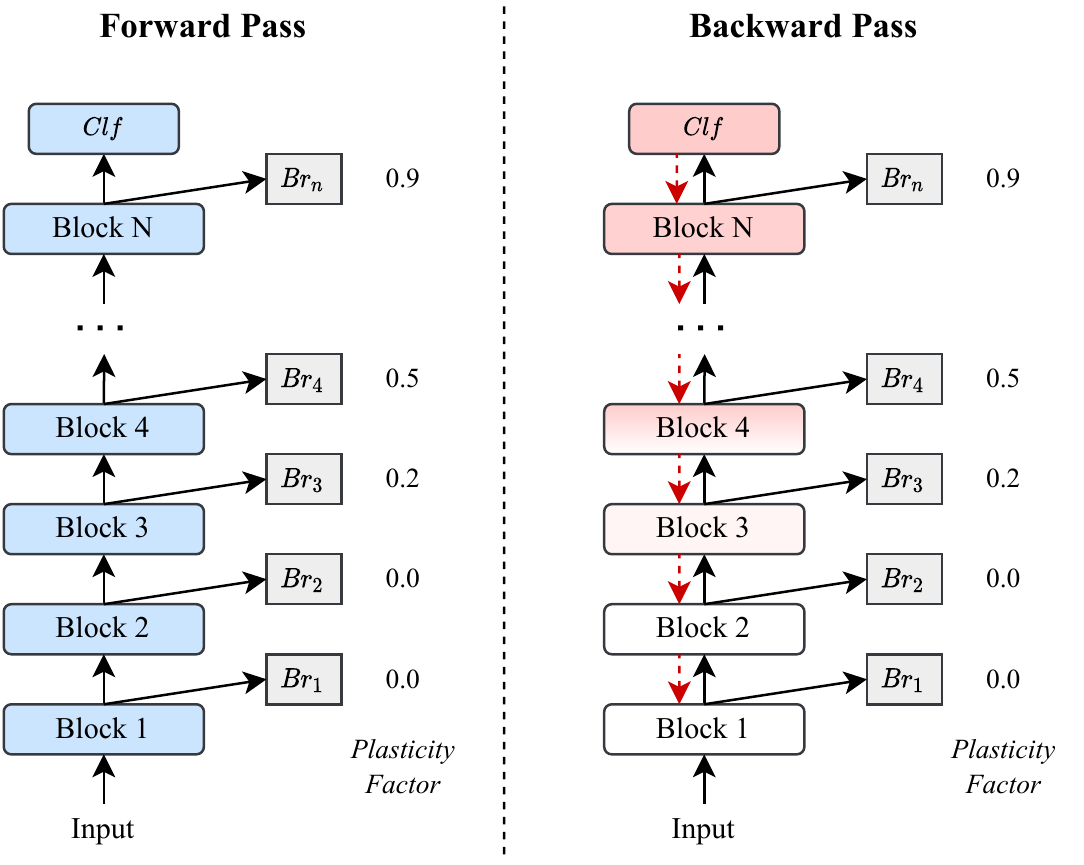}
   \caption{Overview of the method. During the forward step, the backbone processes an example and generates prediction and plasticity factor values for each block (left). During the backward pass, the plasticity factor is used to adjust the final amount of modification each layer will have (right).}
   \label{fig:method}
\end{figure}

In this work, we consider a lifelong (continual) learning setup. Each task $t$ consists of a new data distribution $D^t = (X^t,Y^t)$, where $X^t$ denotes the input instances and $Y^t$ denotes the instance labels. The goal is to train a classification model $ f: X \longrightarrow Y $ using data from a sequence of $T$ tasks: $ D = \{D^1, ..., D^T \} $. Following the Class Incremental Learning setup for CL~\cite{van2019three}, each task is presented sequentially to the model without a task descriptor. Also, in our setup, we only allow each item to be viewed once, such as in online learning scenarios \cite{NEURIPS2019_f8d2e80c, mai2022online}.

A model in this configuration consists of an encoder and a decoder.
The encoder takes an input $x$ and produces a vector representation. The encoder could be any kind of model, for instance, a Transformer \cite{NIPS2017_3f5ee243} for text classification, or a ResNet \cite{he2016deep} for image classification. The decoder is a linear transformation and a softmax layer to predict the class $y$ of an input $x$. Note that because there is no task descriptor, the decoder predicts across all classes.

Next, we explain the proposed method and how it is incorporated into the learning process.

\subsection{Entropy-based Stability-Plasticity (ESP)}

Previous works attempt to solve the stability-plasticity dilemma by slowing down learning on certain weights based on how important they are to previously seen tasks \cite{Kirkpatrick2017, 10.5555/3305890.3306093}.
Although effective, these methods neglect that some layers learn general or task-specific patterns \cite{Zeiler2014,NIPS2014_375c7134,Kovaleva2019,lee2019elsa} and constantly update the weights of the different layers, resulting in interference with the acquired knowledge.
Based on this, ESP addresses the stability-plasticity dilemma using a mechanism that allows the model to decide how much each layer should be updated using an entropy-based criterion.
This method favors the reuse of previous knowledge existing in the layers by means of little or no modification of their weights and the specialization of the layers in a specific task by means of a high modification of their weights.

Under the CL setup mentioned above, an encoder could be seen as a stack of processing blocks.
As shown in \cref{fig:method}, ESP extends each model block $i$ with a side branch layer $Br_i$ to generate a classification of the input $x$:

\begin{equation}
  \hat{y_i} = W_i^2(\sigma(W_i^1 f_i(x)))
  \label{eq:branch_clfs}
\end{equation}

\noindent where $f_i(x)$ is the output of the block $i$, $W_i^1$ and $W_i^2$ are trainable linear layers, and $\sigma$ is an activation function. 

Later, the vector $\hat{y_i}$ is used to compute the entropy of the prediction probability distribution for each block:

\begin{equation}
  E(\hat{y}_i) = \sum \hat{y}_i \log \hat{y}_i 
  \label{eq:entropy}
\end{equation}

Finally, a plasticity factor (PF) is calculated as the complementary entropy value. Note that a $softmax$ function is applied first to find the proportion of entropy corresponding to each block $i$.

\begin{equation}
  PF = (1 - softmax(E(\hat{y})))
  \label{eq:plas_factor}
\end{equation}

This whole process is slightly similar to previous work using branch classifiers and entropy for early exiting \cite{7900006,Xin2020, zhou2020bert,eyzaguirre2022dactbert}. However, in this case, the calculated factors provide the proportion by which the weights of each model layer should be modified.
Intuitively, a high value of PF (low entropy) leads to a high modification of the weights, specializing them for the task. On the other hand, a low value of PF (high entropy) leads to smaller changes in weights, reducing interference and catastrophic forgetting. 

\subsection{Training}
Analogous to EWC and SI, which have an additional step to find the importance of each weight after each new task, ESP needs to update each block's branch layers $Br$ before training a new task.
To do that, we first freeze the encoder and decoder and train only the attached layers with a subset of the training set (e.g., replay set).
We train the layers as a classification problem, where the output of each layer is compared with ground truth using a \textit{Cross Entropy} loss.

\begin{equation}
  Loss_i = CrossEntropy(\hat{y_i},y_i)
  \label{eq:crossentropy}
\end{equation}

Then, to train the backbone (encoder and decoder), the branch layers must be frozen. The reason for freezing the branch layers is to maintain the optimal quality of the decoder classifier. If the branch layers are not frozen, the model layers will no longer be optimized solely for the decoder classifier, which generally worsens its quality. 
% Additionally, branch layer classifications also worsen if all the network is trained together.
Empirically, we found that joint optimization of branch layers and the backbone also leads to worst results.

This training step uses the information provided by the frozen branch layers to self-regulate the weight update.
During the forward pass (\cref{fig:method} left), the model generates an output $y$ and each $Br$ generates the corresponding $PF$. During the backward pass (\cref{fig:method} right), the $PF$ scales the gradient of the corresponding block to control the modifications that the task wants to make to the model. Note the encoder and decoder are optimized with the same loss function of \cref{eq:crossentropy}. 
The training of the model with ESP is summarized in \cref{alg:training}.

\begin{algorithm}[h]
\SetAlgoLined
\textbf{Components:}
\begin{itemize}
    \begin{small}
    \itemsep-0.25em 
    \item $D^t$: Dataset for task $t$.
    \item $F$: Model.
    \item $BC$: Branch Layers.
    \end{small}
\end{itemize}
%\vspace{-2mm}
$BC \leftarrow TrainBC(F, D^t)$ \\
 \For{x,y \KwTo $D^t$}{
    \textcolor{blue}{\textit{\# Forward pass}}

    $PF, \hat{y} \leftarrow F(x)$ \\ 
    
    $\triangledown g \leftarrow Loss(y,\hat{y})$ \\ 
    
    \textcolor{blue}{\textit{\# Backward pass}}

    $\triangledown g_{new} \leftarrow UpdateGrad(PF,\triangledown g)$ \\ 
    
    $F \leftarrow UpdateModel(F,\triangledown g_{new})$ \\
}
\caption{ESP Training Process}
\label{alg:training}
\end{algorithm}

\section{Experiments}
In this paper, we test our approach in the domains of natural language and vision. For a fair and consistent comparison, we use the same CL setup (explained in \cref{method}) and the same baselines for both domains.

\subsection{Baselines}
One of the most reliable approaches to overcoming catastrophic forgetting is \textbf{Replay} \cite{10.1162/neco_a_01433}. It involves storing a subset of previous inputs (e.g., sentences) and mixing them with more recent inputs to update the model. The replay subset is usually a percentage of data randomly taken from the training set of previous tasks. As our baseline, we use a standard replay strategy with commonly used percentages.

To compare our method, we consider several well-known methods that attempt to address the stability-plasticity dilemma to apply to our primary baseline. We provide a brief description of each below:

\begin{enumerate}
    \itemsep-0.25em 
    \item \textbf{Stability}: A method that keeps the encoder weights fixed and trains only the decoder (classifier).
    \item \textbf{Plasticity}: A method with complete freedom to train the encoder and decoder (classifier).
    \item \textbf{Linear Plasticity}: A method that, similarly to ESP, uses a factor to scale the gradient of each block. The factors are linearly spaced between 0 and 1 with respect to the model's number blocks, where 0 is for the first block and 1 for the last one.
    \item \textbf{O-EWC}: Online EWC \cite{schwarz2018progress} introduces a regularization term involving the Fisher information matrix that indicates the importance of each of the parameters to previous tasks.
    \item \textbf{SI}: Synaptic Intelligence \cite{10.5555/3305890.3306093} adjusts the plasticity of the model by regularizing the modification of these important weights with a coefficient.
\end{enumerate}

We also consider the class imbalance issue in our experiments. We train all methods in two scenarios: (1) using ONLY the replay set items, similar to \cite{prabhu2020gdumb} but without using a fixed amount of data, and (2) combining ALL data from the current task with the replay set. In the latter case, there could be a significant class imbalance, but a more considerable amount of data would be available to train the model.

% To verify the dynamism of the plasticity factor, we compare ESP to a static factor. Instead of learning the plasticity factor, we add a linear projection between 0 and 1, where the first layers are the least modified, and the upper layers are the ones with the highest factor. We seek to verify that learning this factor is necessary, and a projection alone is insufficient. We call this baseline ESP-Linear.

\subsection{Natural Language}

\begin{table}[]
\centering
\begin{tabular}{c|ccccc}
\toprule
 & \multicolumn{5}{c}{\textbf{Replay}}       \\
 & \textbf{10\%} & \textbf{20\%} & \textbf{30\%} & \textbf{40\%} & \textbf{50\%} \\ 
\midrule
Stability    & 67.8          & 68.8          & 69.1          & 69.4          & 69.7 \\
Plasticity   & 76.6          & 76.9          & 76.9          & 76.8          & 76.9 \\
Linear Plasticity & 76.8          & 77.1          & 76.9          & 77.0          & 77.1 \\
O-EWC        & 76.8          & 76.8          & 76.9          & 76.9          & 77.1 \\
SI           & 76.7          & 76.6          & 76.9          & 76.9         & 77.2 \\
ESP          & \textbf{76.9}          & \textbf{77.3}          & \textbf{77.1}          & \textbf{77.2}          & \textbf{77.5} \\
\bottomrule
\end{tabular}
\caption{Text classification results using replay set concatenated with ALL the training set.}
\label{tab:bert_all}
\end{table}

\begin{table}[]
\centering
\begin{tabular}{c|ccccc}
\toprule
 & \multicolumn{5}{c}{\textbf{Replay}}       \\
 & \textbf{10\%} & \textbf{20\%} & \textbf{30\%} & \textbf{40\%} & \textbf{50\%} \\ 
\midrule
Stability    & 63.3         & 66.8          & 68.0          & 68.6          & 68.9 \\
Plasticity   & 74.9          & 75.6          & 76.3          & 76.6          & 76.8 \\
Linear Plasticity & 74.7          & 75.8          & 76.2          & 76.6          & 76.9 \\
O-EWC        & 74.3          & 75.7          & 76.2          & 76.5          & 76.3 \\
SI           & 74.6          & 76.0          & 76.2          & 76.6          & 76.9 \\
ESP          & \textbf{75.0}          & \textbf{76.1}          & \textbf{76.6}          & \textbf{76.9}          & \textbf{77.3} \\
\bottomrule
\end{tabular}
\caption{Text classification results using ONLY replay set.}
\label{tab:bert_only}
\end{table}

\subsubsection{Implementation Details}
We use BERT \cite{devlin-etal-2019-bert}, a Transformer-based \cite{NIPS2017_3f5ee243} pre-trained language model, as the encoder. As decoder, following original BERT model, we use the first token (special token [CLS]) of the sequence and a classifier to predict the class. In addition, we use the default BERT vocabulary in our experiments.
We use Adam optimizer with a learning rate of $3e^{-5}$ and a training batch of size 32.

\subsubsection{Datasets}
We use publicly available text classification datasets from \cite{NIPS2015_250cf8b5}: (1) AGNews classification, (2) Yelp sentiment analysis, (3) Amazon sentiment analysis, (4) DBPedia article classification and (5) Yahoo questions and answers categorization. We follow the same data processing described in \cite{NEURIPS2019_f8d2e80c}. In total, we have 575,000 training examples and 38,000 test examples with 33 classes from all datasets. In addition, we use the originally proposed dataset orders:

\begin{enumerate}[(i)]
    \itemsep-0.25em 
    \item Yelp → AGNews → DBPedia → Amazon → Yahoo
    \item DBPedia → Yahoo → AGNews → Amazon → Yelp
    \item Yelp → Yahoo → Amazon → DBpedia → AGNews
    \item AGNews → Yelp → Amazon → Yahoo → DBpedia
\end{enumerate}

\subsubsection{Results}

Our results in the natural language experiments are shown in \cref{tab:bert_all} and \cref{tab:bert_only}. The Plasticity model performs well compared to the Stability version. This was expected because the Stability model limits the flexibility of the model to acquire new knowledge.
%Interestingly, the Linear Plasticity model outperforms the Plasticity model in almost all experiments, supporting the hypothesis that updating the lower layers a little and the high ones a lot leads to better learning.
Interestingly, the Linear Plasticity model outperforms the Plasticity model in almost all experiments, supporting the hypothesis that lower blocks need minor updates, and modifying higher blocks leads to better results.  
On the other hand, we find that O-EWC and SI perform similarly or even worse than the Plasticity model in some cases.
This is because these methods perform better under a setup in which a task id is provided. 

In contrast, ESP outperforms all baselines when trained on all experiments.
We found that the increase in performance is consistent in the ONLY and the ALL scenarios. Overall, ESP achieves an accuracy gain of 0.34 and 0.44 points on average (across all replay percentages) over SI and O-EWC, respectively. 
ESP also exceeds Linear Plasticity, which means that dynamic plasticity factors are useful to avoid forgetting.

% \begin{table}[]
% \centering
% \begin{tabular}{c|rrrrr}
% \toprule
%  & 10\% & 20\% & 30\% & 40\% & 50\% \\
% \midrule
% Enc-Dec & \multicolumn{5}{c}{27.9} \\
% \midrule
% Stability    & 63.3         & 66.8          & 68.0          & 68.6          & 68.9 \\
% Plasticity   & 74.9          & 75.6          & 76.3          & 76.6          & 76.8 \\
% O-EWC        & 74.3          & 75.7          & 76.2          & 76.5          & 76.3 \\
% SI           & 74.6          & 76.0          & 76.2          & 76.6          & 76.9 \\
% ESP         & \textbf{75.0}          & \textbf{76.1}          & \textbf{76.6}          & \textbf{76.9}          & \textbf{77.3} \\
% \bottomrule
% \end{tabular}
% \caption{Text classification results for scenario ONLY. \% represents the percentage of replay.}
% \label{tab:bert_only}
% \end{table}

\subsection{Vision}

\begin{table}[]
\centering
\begin{tabular}{c|ccccc}
\toprule
 & \multicolumn{5}{c}{\textbf{Replay}}       \\
 & \textbf{1\%} & \textbf{2\%} & \textbf{3\%} & \textbf{4\%} & \textbf{5\%} \\ 
\midrule
%Enc-Dec & \multicolumn{5}{c}{17.2} \\
%\midrule
Stability  & 15.9          & 17.5          & 18.0          & 20.3          & 19.9 \\
Plasticity & 20.1          & 29.1          & 36.6          & 40.1          & 42.3 \\
Linear Plasticity & 20.3          & 26.1          & 28.9          & 34.6          & 39.7 \\
O-EWC      & 16.5          & 19.7          & 25.4          & 29.3          & 34.6 \\
SI         & \textbf{23.0} & 29.9          & 38.8          & \textbf{40.4} & \textbf{43.8} \\
ESP        & 21.8          & \textbf{31.8} & \textbf{39.1} & 38.6          & 40.6 \\
\bottomrule
\end{tabular}
\caption{Image classification results using replay set concatenated with ALL the training set.}
\label{tab:cifar10_pre_trained_all}
\end{table}

\begin{table}[]
\centering
\begin{tabular}{c|ccccc}
\toprule
 & \multicolumn{5}{c}{\textbf{Replay}}       \\
 & \textbf{1\%} & \textbf{2\%} & \textbf{3\%} & \textbf{4\%} & \textbf{5\%} \\ 
\midrule
%Enc-Dec & \multicolumn{5}{c}{17.2} \\
%\midrule
Stability  & 15.8          & 22.9          & 26.3          & 28.3          & 29.0 \\
Plasticity & 28.0          & 37.3          & 42.6          & 46.6          & \textbf{50.3} \\
Linear Plasticity & 24.6          & 32.6          & 39.6          & 43.6          & 45.5 \\
O-EWC      & \textbf{29.2} & 37.8          & 40.1          & 45.7          & 47.5 \\
SI         & 28.9          & \textbf{38.5} & \textbf{44.0} & \textbf{47.1} & 49.6 \\
ESP        & 26.0          & 36.1          & 41.2          & 46.2          & 48.5 \\
\bottomrule
\end{tabular}
\caption{Image classification results using ONLY replay set.}
\label{tab:cifar10_pre_trained_only}
\end{table}

\subsubsection{Implementation Details}
For the visual experiments, as encoder we use a pre-trained ResNet-18 \cite{he2016deep}, and following previous works, a linear classifier for the decoder. Instead of using a branch classifier for each layer, here we use a branch classifier for each block of ResNet. The output of each block is reduced to one element per channel by averaging the values of the activation maps, this vector goes through the branch functions to find the scores.

We use SGD as our optimizer, using a learning rate of $1e^{-3}$ and momentum factor equal to $0.9$. We run all of our experiments using a batch size of 32. For EWC and SI, we try different values of regularization coefficient, at the end we use 2000 for EWC and 0.1 for SI.

\subsubsection{Datasets}

Following previous works \cite{10.5555/3305890.3306093}, we use CIFAR10 \cite{krizhevsky2009learning} equally divided into 5 tasks. We use the implementation from Avalanche \cite{lomonaco2021avalanche} to generate the different sequence. We run each experiments 3 times with different seeds and we average the results.

\begin{figure*}[!htp]
  \centering
  \begin{subfigure}{0.48\linewidth}
    \includegraphics[width=0.99\linewidth]{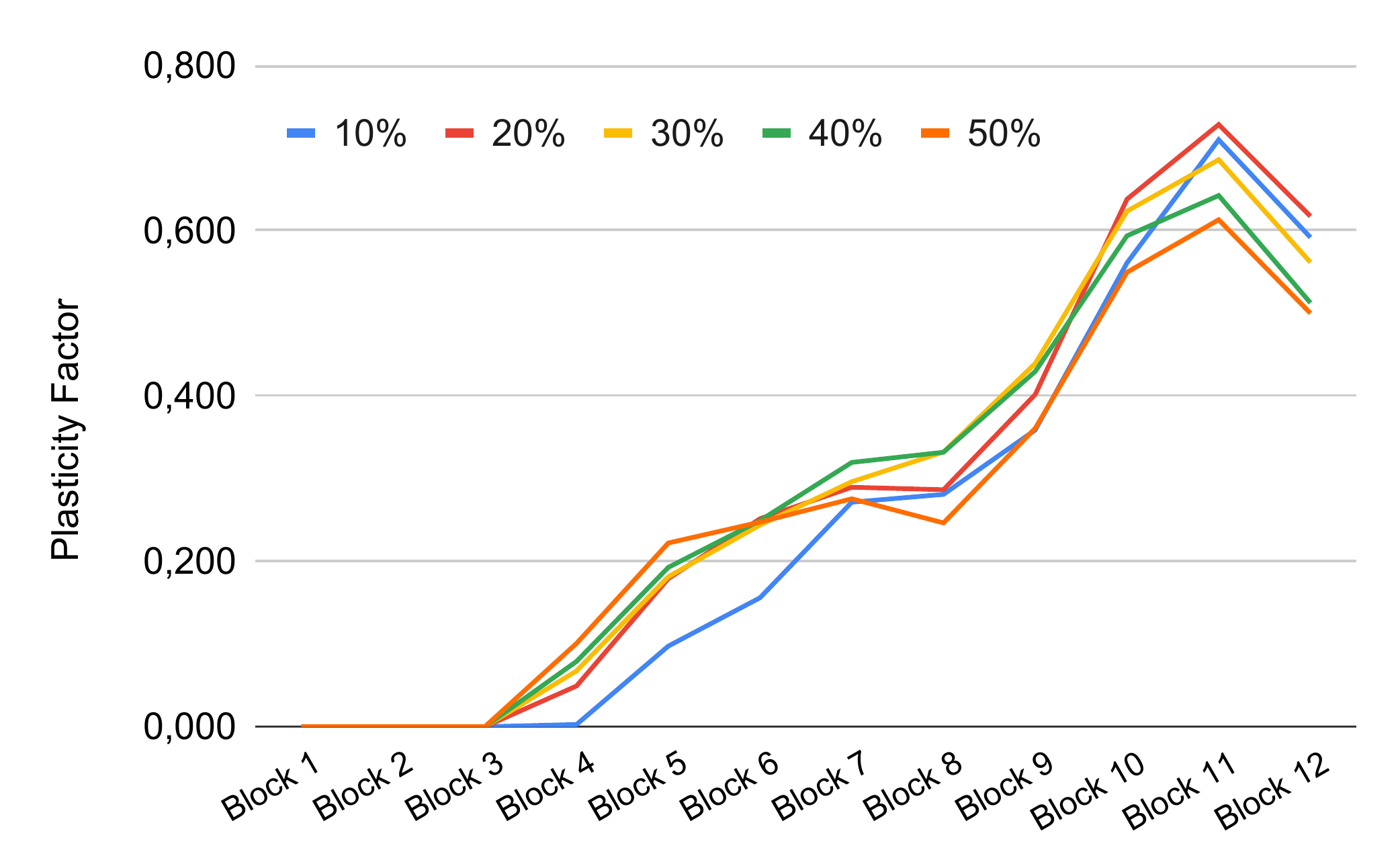}
    \caption{Natural language results with BERT base model.}
    \label{fig:short-a}
  \end{subfigure}
  \hfill
  \begin{subfigure}{0.48\linewidth}
    \includegraphics[width=0.99\linewidth]{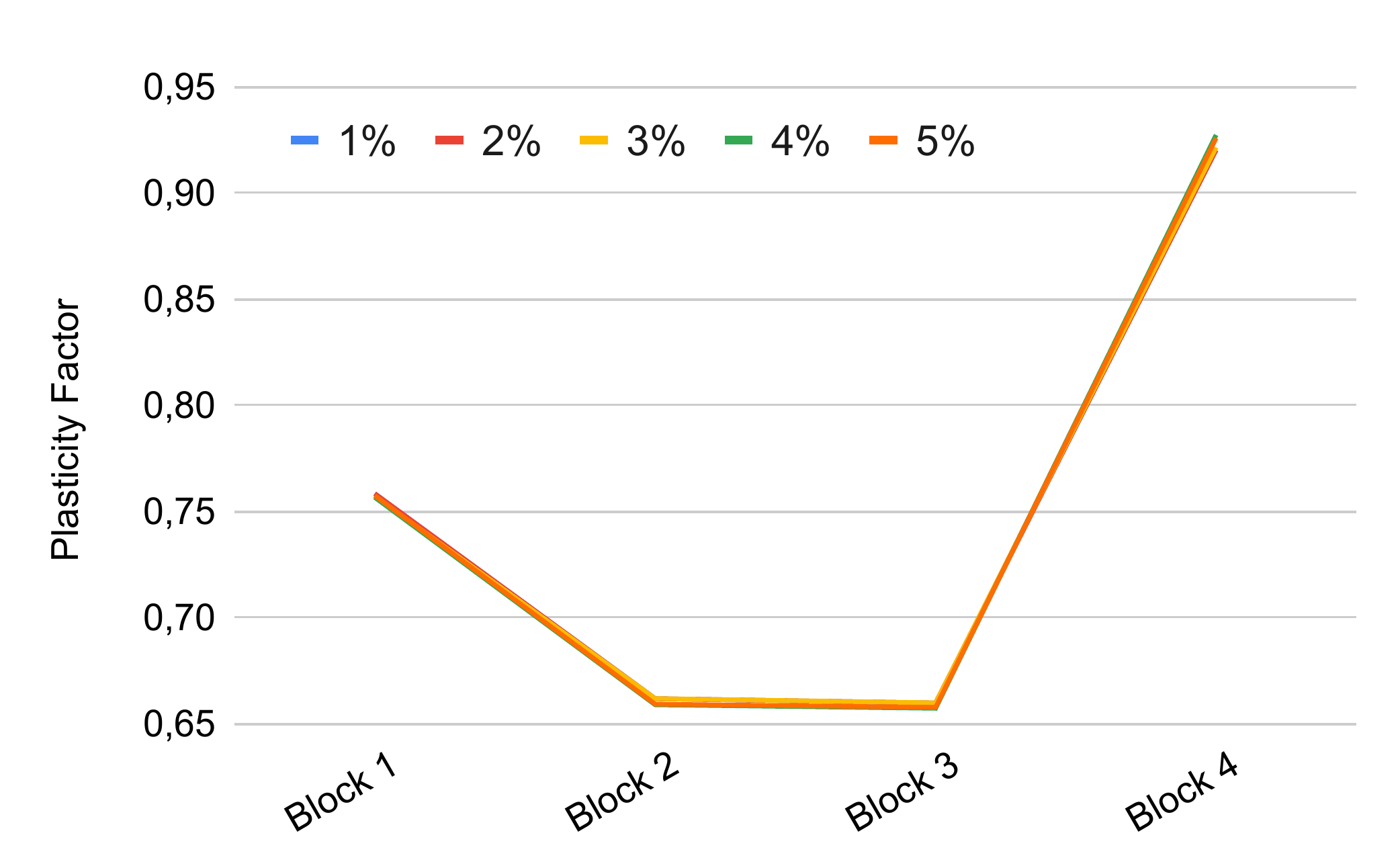}
    \caption{Vision results with ResNet-18 model.}
    \label{fig:short-b}
  \end{subfigure}
  \caption{Average plasticity factor per block across all tasks. \% represents the percentage amount of replay.}
  \label{fig:plasticityfactors}
\end{figure*}

\subsubsection{Results}

Similar to the natural language experiments, we use a pre-trained model, specifically a ResNet pre-trained on ImageNet. However, a big difference between text and images is the number of elements saved in the replay set. As images weigh more than phrases, we do experiments saving between 1 and 5\% of previous tasks. 

Results shown in \cref{tab:cifar10_pre_trained_all} and \cref{tab:cifar10_pre_trained_only}, shows that regularization methods have better results than Stability, Plasticity and Linear Plasticity.  
% Updating only the linear layer is not enough to achieve a good classification since the distribution of CIFAR10 is different from that of ImageNet. 
On the other hand, modifying the weights without a constraint does not achieve good results either. In general, there is no clear advantage between the regularization methods in either of the two scenarios. 
% Different from the natural language experiments, Linear Plasticity obtains a lower accuracy than Plasticity, reinforcing the idea of modifying the first layers is necessary due of the difference between the ImageNet and CIFAR10.
We hypothesize the results diverge from the natural language experiments for two reasons: The first is the difference in the number of blocks, indicating that ESP may take advantage of deeper networks. The second is because there is a difference in the input distribution between ImageNet and CIFAR10, we expand this hypothesis a bit more in \cref{section:further}.

The motivation behind regularization methods is to reduce the plasticity of the model to prevent forgetting. By minimizing the modification of relevant weights in the future training process, these methods force future tasks to reuse knowledge even if those patterns are irrelevant or hurtful to new tasks. For this reason, we believe it is essential for these methods to learn representations that may be useful across tasks. For example, if the model weights are too specific for a task and we freeze all layers, the model would not find a correct classification.

Given the above, we believe that one reason why regularization methods do not perform well in Class Incremental scenarios is the inability to find good representations and thus the need to start from a pre-trained model. To prove this hypothesis and compare our results, we change the pre-trained ResNet-18 to one initialize randomly.
The results show that none of the three regularizing methods has good results, being outperformed by the Plasticity method in almost all replay percentages. The advantage of this method is that it has complete flexibility to adjust the weights. This advantage leads to the weights to learn new representations, not tied to representations particular to the previous tasks. It is of little use to reduce the modification of past relevant weights if they can not be reused for future tasks.

\subsection{Further Analysis}
\label{section:further}

This section discusses the plasticity factors resulting from our experiments in both domains. \cref{fig:plasticityfactors} shows the average of the plasticity factors for our experiments on natural language (\cref{fig:short-a}) and vision (\cref{fig:short-b}).

We use a pre-trained BERT, a 12 block model for natural language. Interestingly, the plasticity factor of the lower layers (1 to 4) is 0, and the factor constantly increases for the upper layers. Which means null modification in the lower layers and high modification in the upper layers.
Our method leverages existing general knowledge in lower layers, which is general knowledge, while updating upper layers with task-specific knowledge.

Regarding vision, we use a pre-trained ResNet-18, a four-block model. Here our method behaves slightly similar to the natural language model, where the block with higher modifications is the last one. However, unlike the natural language results, the first block has a higher plasticity factor than blocks 2 and 3. This result may be due to the data used for pre-train the model. 
The BERT model is pre-trained on a massive corpus of different domains and topics, promoting the lower layers to be general for any task. On the other hand, ResNet-18 was pre-trained on ImageNet, which has much higher resolution images than CIFAR10. The basic patterns are expected to be different between both datasets, explaining the high plasticity factor of the first block.
% In language, words have a similar meaning regardless of the context in which they are used (general knowledge) and only affect task-specific patterns. 
% In images, the model is pre-trained with ImageNet, which has much larger images than CIFAR10. The basic patterns are expected to be different between both datasets, which would explain the significant variability of the first block.

In general, the plasticity factors for natural language and vision are higher at the last layers. However, no one reaches complete plasticity, which means those layers retain some specialization acquired in previous tasks.
% Also, \cref{fig:plasticityfactors} shows that the graphs per different replays are almost the same, and this is because the plasticity of the network only depends on the current input.
Also, \cref{fig:plasticityfactors} it shows that different amounts of percentages of replay sets have similar results, indicating that the plasticity of the network mainly depends on the current input.

\begin{figure}[t]
  \centering
   \includegraphics[width=0.99\linewidth]{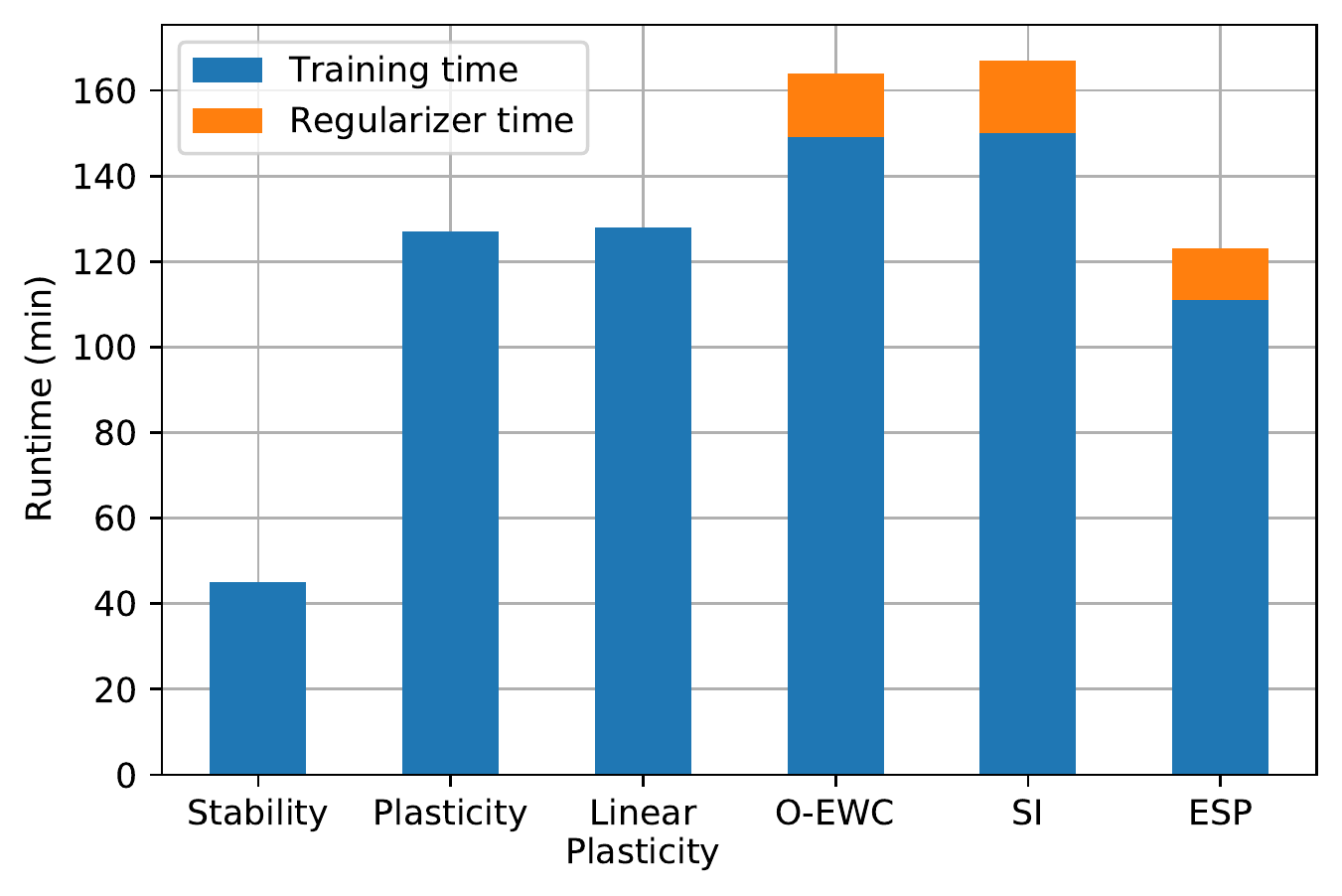}
   \caption{Training time comparison (in minutes) for all models with a replay of 20\%. The blue color corresponds to the training time of the backbone and the orange color corresponds to the additional time that the regularizer takes.}
   \label{fig:runtime}
\end{figure}

Finally, we argue that ESP could be computationally efficient for cases like natural language experiments. \cref{fig:runtime} shows the training runtime of ESP and baselines under the 20\% replay setup. We use a GPU NVIDIA GeForce RTX 3090 and a CPU AMD EPYC 7502 for these experiments. The Stability model is the more efficient, with $\sim$45 minutes, because it only updates the decoder layer. The Plasticity and Linear Plasticity model take $\sim$127 minutes because the entire backbone is trained. 
On the other hand, O-EWC and SI methods take a total time of $\sim$164 and $\sim$167 minutes. For a fair comparison, we divide the total execution time into the backbone training time (blue) and the regularizer time (orange) because both models include an additional process to calculate the importance of the weights. O-EWC takes $\sim$149 minutes of backbone training time and $\sim$15 minutes to find the importance of each weight. SI takes $\sim$150 minutes of backbone training time and $\sim$17 minutes of regularization time. Note that the backbone training time of these methods is superior to the plasticity model because they include an additional calculation loss based on the importance of the previously calculated weights.

Concerning our method, ESP finishes its training in $\sim$123 minutes. Analogous to O-EWC and SI, ESP has an additional process to tune the branch layer, which takes $\sim$12 minutes. This time is similar compared to O-EWC and SI regularizer time. However, if we compare the training time of the spine, ESP is remarkably efficient. ESP takes $\sim$111 minutes, which is less than other methods, including the Plasticity model.
It happens because ESP sometimes forces the model not to update some layers, allowing those layers to be frozen on the fly, resulting in decreased training time.

\section{Conclusion}
In this paper, we introduced ESP, a method based on an entropy-based criterion to decide how much a model has to modify the weights of each of its layers. ESP augments each block of an encoder with branch layers that computes an entropy-based plasticity factor used to update layer weights dynamically. Our experiments in the natural language and vision domains show the effectiveness of our model in leveraging prior knowledge by not updating lower layers and specializing other layers by updating higher layers. In addition, we show that in the case of the natural language model, our method promotes computational efficiency since it forces not to update some layers.

Among the ideas for future work, we consider testing the hypothesis that ESP works better on networks with more blocks than a Resnet-18, such as vision Transformers. Also, we would like to extend ESP to an utterly online scenario.

\section*{Acknowledgement}
We thanks the reviewers for their positive remarks and some valuable suggestions. This work was supported by the European Research Council Advanced Grant 788506 and the National Center for Artificial Intelligence CENIA FB210017, Basal ANID.

%%%%%%%%% REFERENCES
{\small
\bibliographystyle{ieee_fullname}
\bibliography{egbib}
}

\end{document}